\documentclass[letterpaper]{article} 
\usepackage{aaai21}  
\usepackage{times}  
\usepackage{helvet} 
\usepackage{courier}  
\usepackage[hyphens]{url}  
\usepackage{graphicx} 
\usepackage{natbib}  
\usepackage{caption} 
\usepackage{amssymb}
\usepackage{amsmath}
\usepackage{tabularx}
\usepackage{booktabs}
\usepackage{subfig}
\usepackage{pifont}
\usepackage[switch]{lineno}

\title{Smoother Network Tuning and Interpolation \\ for Continuous-level Image Processing}
\author{
	Hyeongmin Lee\textsuperscript{$\star$\rm 1}, Taeoh Kim\textsuperscript{$\star$\rm 1},\\
	Hanbin Son\textsuperscript{\rm 1},
	Sangwook Baek\textsuperscript{\rm 2},
	Minsu Cheon\textsuperscript{\rm 3},
	and Sangyoun Lee\textsuperscript{$\dagger$\rm 1} \\
}
\affiliations{
	\textsuperscript{\rm 1}Yonsei University, Seoul, Korea \
	\textsuperscript{\rm 2}Samsung Research, Seoul, Korea \
	\textsuperscript{\rm 3}MathWorks \

	\{minimonia, kto, hbson, syleee\}@yonsei.ac.kr, sw123.baek@samsung.com, mcheon@mathworks.com
	
}

\begin{document}

\maketitle

\let\thefootnote\relax\footnote{{\textsuperscript{$\star$}Equal Contribution}}
\let\thefootnote\relax\footnote{{\textsuperscript{$\dagger$}Corresponding Author}}

\begin{abstract}
In Convolutional Neural Network (CNN) based image processing, most studies propose networks that are optimized to single-level (or single-objective); thus, they underperform on other levels and must be retrained for delivery of optimal performance. Using multiple models to cover multiple levels involves very high computational costs. To solve these problems, recent approaches train networks on two different levels and propose their own interpolation methods to enable arbitrary intermediate levels. However, many of them fail to generalize or 
have certain side effects in practical usage. In this paper, we define these frameworks as \textit{network tuning and interpolation} and propose a novel module for continuous-level learning, called Filter Transition Network (FTN). This module is a structurally smoother module than existing ones. Therefore, the frameworks with FTN generalize well across various tasks and networks and cause fewer undesirable side effects. For stable learning of FTN, we additionally propose a method to initialize non-linear neural network layers with identity mappings. Extensive results for various image processing tasks indicate that the performance of FTN is comparable in multiple continuous levels, and is significantly smoother and lighter than that of other frameworks.

\end{abstract}

\section{Introduction}
\label{intro}

Image processing algorithms have various objectives that can include a combination of objective functions or a pair of target level-specific training datasets.
For example, in restoration tasks such as denoising, there is an optimal level for each input whose noise level is unknown, and in image synthesis, balancing fidelity and naturalness \cite{pdt2018} depends on target applications.
In style transfer, the user hopes to control various styles and stylization strengths continuously.

\begin{figure}[!t]
	\begin{center}
		\subfloat[\small{Multi-task Learning}]
		{\includegraphics[width=0.495\linewidth]{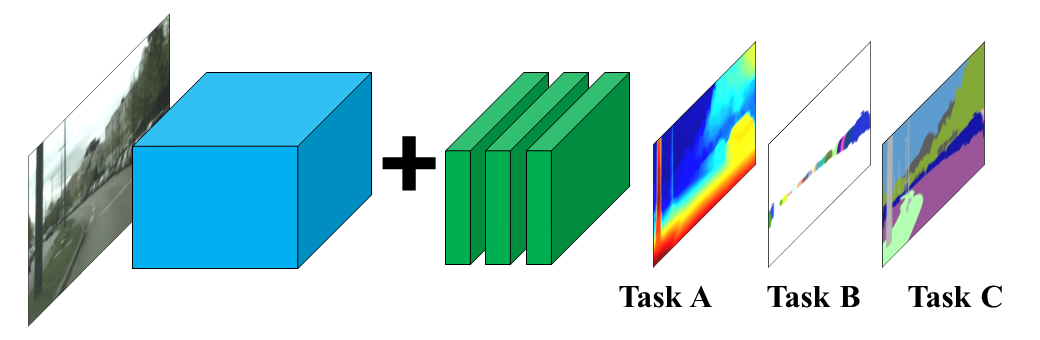}}
		\hfill
		\subfloat[\small{Multi-level Learning}]
		{\includegraphics[width=0.495\linewidth]{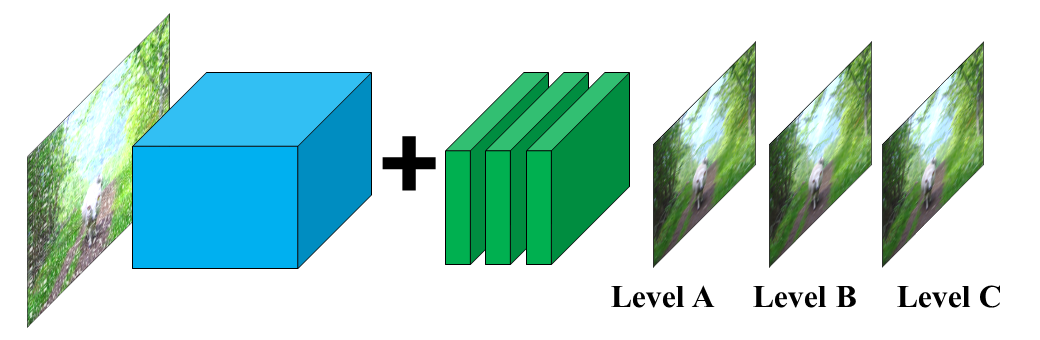}}\,\\[0.2ex]
		\subfloat[\small{Continuous-level Learning}]
		{\includegraphics[width=0.55\linewidth]{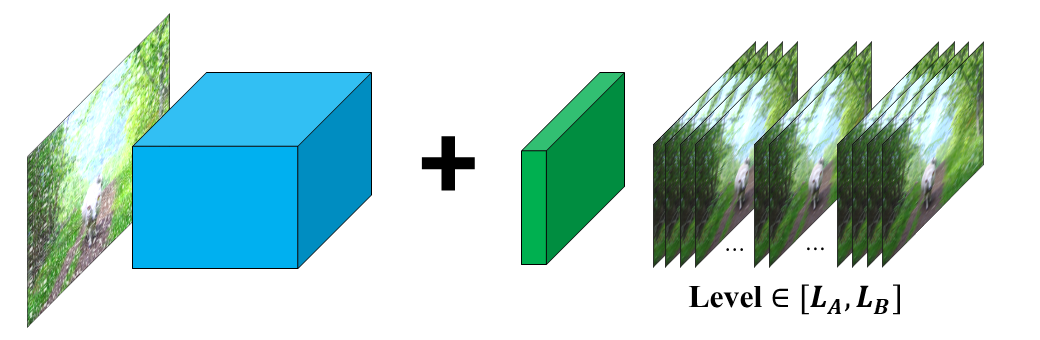}}\\[-1.3ex]
		\caption{{Comparison of multi-task learning, multi-level learning, and continuous-level learning. Every task or level shares a main network ({blue}) and introduces an additional branch ({green}) for task- or level-specific optimization}}
		\label{fig:cmll}
	\end{center}
\end{figure}

However, most image processing deep networks are trained and optimized for single-level. 
In this paper, the word \textbf{\textit{level}} can be one of the following examples: a target noise level (standard deviation of Gaussian noise or quality factor of JPEG), a specific combination of objective functions to optimize, or a target style for style transfer.
If we want to handle $N$ multiple levels, we must train $N$ different models or exploit the structure of multi-task learning \cite{edsr2017} (Fig. \ref{fig:cmll} (b)), which is inefficient when $N$ increases.
In addition, in many image processing tasks, levels can be continuous.  
Therefore, designing a network for a continuous-level in an efficient way is a very practical issue. Fig. \ref{fig:cmll} describes the differences among multi-task learning, multi-level learning, and continuous-level learning. Compared to multi-task learning, multi-level learning solves single-task and multiple discrete-level problems. Continuous-Level Learning (CLL) is an extension of multi-level learning whose levels are continuous between two levels, which is a general issue in image processing tasks.

To deal with CLL problems, several frameworks have been proposed~\cite{adafm2019,dynamicnet2019,cfsnet2019,dni2019,esrgan2018}, and they have the following steps in common. In the training phase, they train their CNN network twice for each level. The first-training is similar to the general network training method. During the second-training, some parameters that were optimized in the first-training are fixed, and the other parameters are fine-tuned or some additional modules are trained for the second level. In the test phase, they make their networks available at any intermediate level with their own interpolation methods. We define these frameworks as \textit{network tuning and interpolation}. These steps are derived from observations in~\cite{adafm2019,dni2019}. They show that the fine-tuned filters are similar to those of the original filters, which makes the interpolation space between filters meaningful.

Although various \textit{network tuning and interpolation} frameworks have been proposed, deeper analysis and stable generalization for practical usage are required. We define three aspects to better utilize CLL algorithms. 
The first is \textbf{\textit{adaptation and interpolation performance}}. After the second training, its performance might be lower than that of the one trained only for the second level, because it contains parameters for both levels. Therefore, CLL frameworks have to be flexible in order to adapt to new levels. In addition, even though the network works well on the two trained levels, it might not work for the other intermediate levels. 
Therefore, it is also important for the networks to maintain high performance and reasonable outputs for intermediate levels. We can measure them using the metrics for each task at arbitrary intermediate unseen continuous levels.
The second one is \textbf{\textit{smoothness}} for practical usage scenarios. During the experiment, we find that certain artifacts and unintended behaviors are caused by some CLL methods. Exhibiting stable performance across tasks and networks, not producing undesirable artifacts, and operating with interpretable control parameters are very important for real-world usage, and we define these aspects collectively as \textit{smoothness}. 
The last one is \textbf{\textit{efficiency}}. 
Because one of the main objectives of CLL is to use a single network instead of using multiple networks trained for each level, requiring too large memory and computational resources is not practical for real-world applications.


Most of the prior approaches have limitations in terms of the above three aspects. AdaFM~\cite{adafm2019} introduces a tuning layer called the feature modification layer, which is a simple linear transition block (depth-wise convolution). However because AdaFM is originally proposed for image restoration tasks only, linearity reduces the flexibility of adaptation. Therefore, it is not appropriate for more complex tasks such as the perception-distortion (PD)  trade-off in restoration or style transfer.
Deep Network Interpolation (DNI)~\cite{dni2019,esrgan2018} interpolates all parameters in two distinct networks trained for each level to increase flexibility. One is the version trained from the initial state, and the other is the version fine-tuned starting from the first one. However, fine-tuning the network without any constraint cannot consider the initial level, which might lead to a degraded performance at intermediate levels.
In fact, from the experiments, DNI has limitations on the smoothness conditions. 
DNI also requires extra memory to save temporary network parameters and a third interpolated network for the inference.
CFS-Net~\cite{cfsnet2019} and Dynamic~Net~\cite{dynamicnet2019} propose frameworks that use additional tuning branches to interpolate the feature maps, not the model parameters.
However, tuning branches require large memory and heavy computations up to twice the baseline networks. 
Training two branches independently can cause over-smoothing artifacts because each branch cannot consider each other. This side effect will be discussed later.

In this paper, we propose a novel smoother \textit{network tuning and interpolation} method using a \textit{Filter Transition Network (FTN)} that take CNN filters as input and learns the transitions between levels. Because FTN is a non-linear module, networks can better adapt to any new level than the linear one. Therefore, it can cover general image processing tasks from simple image denoising to complex stylization tasks. FTN \textit{transforms} the filters of the main network via other learnable networks, and we can control the flexibility of transformation by restricting the learnable parameters for smooth and stable interpolation.
For efficiency, from the motivations in~\cite{adafm2019,dni2019}, FTN directly changes filters to be data-agnostic. In other words, because FTN takes filters as input instead of feature maps, the computational complexity does not increase when the input images increase in size. In addition, randomly initialized FTN makes the training process unstable because it directly changes the model parameters. To solve this problem, we propose a method to initialize multiple nonlinear layers to be identity mappings.

In summary, the proposed framework has the following contributions:

\begin{itemize}
	\item We propose a smoother network tuning and interpolation method for CLL using the FTN, which is structurally smoother than the other frameworks. In addition, for the stable learning of FTN, we propose a new initialization method that makes non-linear network layers be identity mapping.
	\item We define and point out the smoothness conditions for CLL which is important for practical applications such as generalization across tasks and networks, color artifacts, and interpretable control parameters.
	\item Our method is comparable in adaptation and interpolation performance on multiple imaging levels, significantly smoother in practice, and efficient in both memory and computational complexity.
\end{itemize}

\section{Related Work}

In this section, we summarize the CLL problems in image processing tasks which will be experimented in this paper.

\begin{figure*}[!t]
	\begin{center}
		\includegraphics[width=0.85\linewidth]{"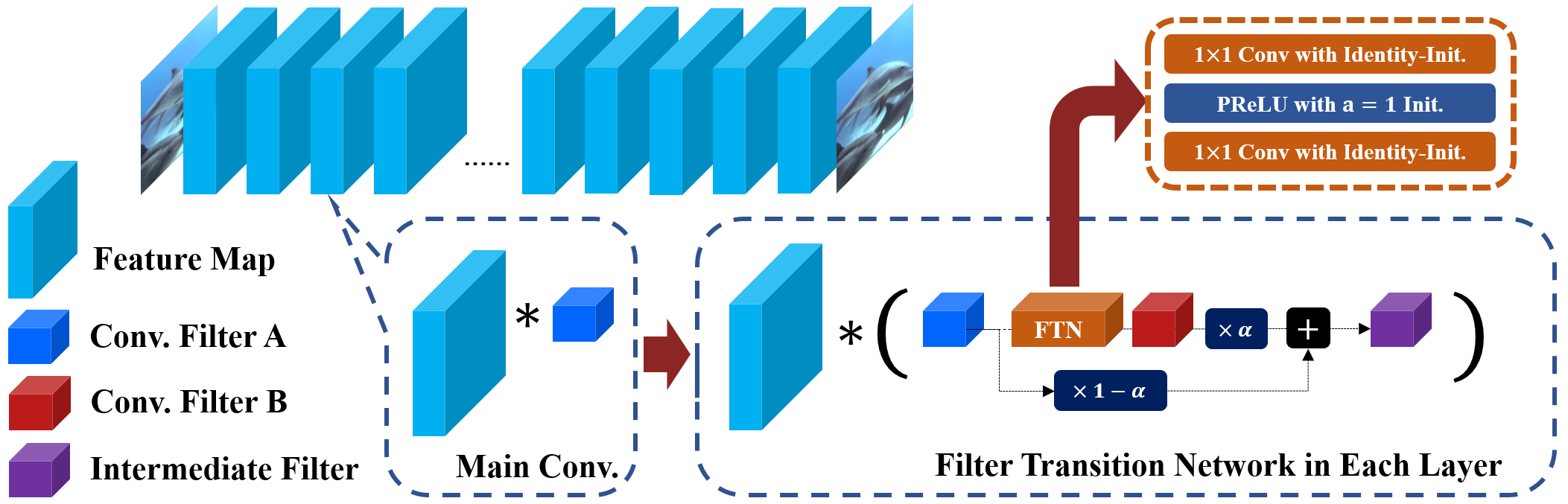}
	\end{center}
	\caption{{Network architecture of the proposed Filter Transition Network (FTN) when adapted in arbitrary main convolutional networks. Filter of main network ({blue}) is transformed via FTN for other levels ({red}). In inference phase, interpolated filter ({purple}) is used for intermediate levels}}
	\label{fig:network}
\end{figure*}

\subsubsection{Image Restoration.}
CNN-based image restoration has shown great performance improvements over handcrafted algorithms. After shallow networks, \cite{arcnn2015,srcnn2015}, some works stacked deeper layers, exploiting the advantages of residual skip-connection \cite{vdsr2016,dncnn2017}. Following the evolution of image recognition networks, restoration networks have focused on the coarse-to-fine scheme \cite{lapsrn2017}, dense connections \cite{rdn2018}, attentions \cite{rcan2018} and non-local networks \cite{nlrn2018}.
However, most networks are trained and optimized for a single level such as the Gaussian noise level in denoising, quality factor in JPEG compression artifact removal, and super-resolution scale in single-image super-resolution.
If the levels of training and test do not match, then the optimal restoration performance cannot be achieved.
To deal with this limitation, \cite{mildenhall2018burst,zhang2018ffdnet} proposed multiple noise-level training with a noise-level map, or noise estimation network \cite{guo2019toward} can be a solution. However, the user cannot control the test phase for better personalization~(\textit{e.g.} level of smoothing).

\subsubsection{The Perception-Distortion Trade-off.}
In comparison with the general approach that attempts to reduce pixel-error with the ground truth, some works \cite{argan2017,srgan2017,esrgan2018} attempted to produce more natural images using the generative power of GANs \cite{gan2014,cgan,dcgan}.
They used a combined loss of the fidelity and adversarial terms and then obtained better perceptual quality.
However, when a more adversarial loss is used, worse fidelity with the ground truth occurs because of the perception-distortion (PD) trade-off \cite{pdt2018}.
In \cite{pdt2018}, they proposed evaluating the restoration performance via a PD-plane considering the balance between fidelity and naturalness. However, the network must be retrained on another loss function to draw a continuous PD-function, which is very time-consuming.

\subsubsection{Style Transfer.}
With regard to image style transfer, Gatys~\emph{et~al.}~\cite{styletransfer2015} proposed a combination of content loss and style loss, and optimized content images via pre-trained feature extraction networks. Johnson~\emph{et~al.}~\cite{perceptual2016} made it possible to operate in a feed-forward manner using an image transformation network. However, a network trained on a single objective cannot control the balance between content and style and cannot handle continuous styles when it is trained on a single style.
Even though \cite{gatys2017controlling} can control several factors in the training phase and arbitrary (Zero-shot) style transfer such as \cite{huang2017arbitrary,sheng2018avatar} can handle infinite styles using adaptive instance normalization or style decorator, none of these can control continuous objectives (losses) during the test phase.

\section{Proposed Approach}
\subsection{Filter Transition Networks}
\label{defconv}

The general concept of our module is the same as that of the prior CLL frameworks, \textit{network tuning and interpolation} which was described in the introduction. Our overall framework is detailed in Fig. \ref{fig:network}. Our FTN module in an arbitrary convolutional layer can be described as
\begin{equation}
\textbf{Y}_{i} = \textbf{X}_{i} * (\textbf{f}^{(1)}_{i} \times (1-\alpha) + FTN(\textbf{f}^{(1)}_{i}) \times \alpha )
\label{eq:ftn}
\end{equation}

\noindent where $\textbf{X}_{i}$ and $\textbf{Y}_{i}$ are the input and output of the $i$-th convolution layer, $\textbf{f}^{(1)}_{i}$ is the corresponding kernel, $\alpha$ is the control parameter between $[0, 1]$, and $*$ is the convolution operation. 
A remarkable difference from the existing frameworks is that FTN directly takes network kernels as inputs, instead of images or feature maps. It can be viewed as one variation of hyper-networks~\cite{hypernetworks} which takes and predicts network parameters. Our design goal is \textit{complete adaptation with minimum filter change}. In other words, it is essential to keep our filters as similar as possible to the original filters while adapting another level well. We assume that this can increase \textit{smoothness} and empirical results for this assumption are described in the \textit{smoothness} section of the experimental results.

The FTN consists of two $1\times1$  convolutions with a $G$ grouped convolution  \cite{alexnet,resnext}, PReLU~\cite{prelu2015} activation functions, and skip-connection with weighted sum.
First, we train the main convolutional filter for the initial level with $\alpha=0$. Then, we freeze the main network and train the FTN only for the second level with $\alpha=1$, which breaks skip-connection. Next, the FTN learns the task transition itself. To that end, the FTN approximates kernels of the second level, as in $FTN(\textbf{f}^{(1)}_{i}) \approx \textbf{f}^{(2)}_{i}$, where $\textbf{f}^{(2)}_{i}$ is an optimal kernel for the second level. In the inference phase, we can interpolate between two kernels (levels) by choosing $\alpha$ in the 0-1 range, and Eq. (\ref{eq:ftn}). Consequently, the FTN implicitly learns continuous transitions between levels, and $\alpha$ represents the amount of filter transition towards the second level.

Group convolution can reduce the number of parameters in a network. If the number of groups is increased, the degrees of freedom to change the original filters decrease. We use this relationship to improve smoothness at the expense of adaptation and interpolation performance. 
$1\times1$ convolution is used because: 1) it is lightweight, and 2) padding is not required. Because the input size of the FTN is quite small (usually $3 \times 3 \times C$), padding can be critical to each layer. 

\subsection{Initialization of FTN}

During the second-training, since we set $\alpha=1$, each convolution layer can be formulated as follows.
\begin{equation}
\textbf{Y}_{i} = \textbf{X}_{i} * (FTN(\textbf{f}^{(1)}_{i}))
\label{eq:ftn_init}
\end{equation}

However, when we initialize FTN using general methods such as \cite{xavier2010,prelu2015}, which will predict random filters from $\textbf{f}^{(1)}_{i}$, the training cannot start from the first level ($FTN(\textbf{f}^{(1)}_{i}) \neq \textbf{f}^{(1)}_{i}$). These types of initialization make the training very unstable if special precautions are not taken. In our framework, every convolution and activation function is initialized as an identity function. Convolutions can easily become identities~\cite{adafm2019}. For activation functions, we use PReLU \cite{prelu2015} with an initial negative slope $a=1$, which will be learned through training.

\section{Experiments}

\subsection{Experimental Settings}
\label{setup}

\begin{table}[!b]
	\centering
	\caption{{\textbf{Ablation study for structures of FTN.} Average PSNR (dB) on CBSD68 denoising test dataset. Unseen noise levels are denoted with 
			*. The baseline network is \textbf{AdaFM-Net}. The best results are \textbf{bold-faced}.}}
	\resizebox{0.75\linewidth}{!}{
		\begin{tabular}{cc|ccccc}
			\toprule
			& Noise Level $\sigma$ & 20  & 30*  & 40*  & 50  \\  \midrule	
			& FTN  & \textbf{32.44}  & \textbf{30.18}  & \textbf{28.90}  & \textbf{28.04} \\
			& FTN-deeper & \textbf{32.44}  & 30.06  & 28.81  & 28.03 \\
			& FTN-spatial & \textbf{32.44}  & 30.16  & 28.88  & \textbf{28.04} \\ \bottomrule	
	\end{tabular}}
	\label{tb:dn_ablation}
\end{table}

\subsubsection{Baselines.}
To understand recent \textit{network tuning and interpolation} frameworks for CLL, we evaluate FTNs against DNI \cite{dni2019}, AdaFM \cite{adafm2019}, CFSNet \cite{cfsnet2019}, and Dynamic-Net \cite{dynamicnet2019} on four general image processing tasks. We add a tuning layer of FTNs into every convolution, and the same for AdaFM \cite{adafm2019} except the last layer to prevent boundary artifacts. We add a ResBlock-wise (or DenseBlock-wise) tuning branch for CFSNet \cite{cfsnet2019}. For a fair comparison, the main networks are identical and shared across frameworks, and every hyper-parameter is identical except for the tuning layers of each framework. More detailed configurations are described in the supplementary material.

\begin{figure}[!t]
	\begin{center}
		\captionsetup[subfigure]{labelformat=empty}
		\rotatebox{90}{\makebox[13mm][c]{\small{\textsc{Style A}}}}
		\subfloat
		{\includegraphics[height=0.157\linewidth]{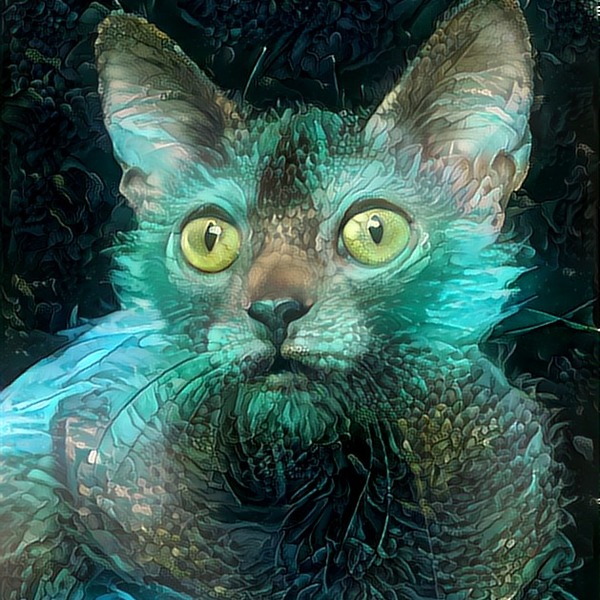}}\
		\hfill
		\rotatebox{90}{\makebox[13mm][c]{\small{\textsc{Content}}}}
		\subfloat
		{\includegraphics[height=0.157\linewidth]{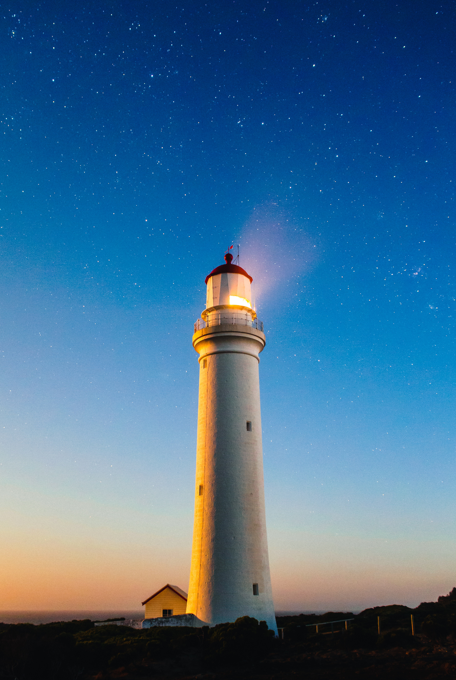}}\
		\hfill
		\rotatebox{90}{\makebox[13mm][c]{\small{\textsc{Style B}}}}
		\subfloat
		{\includegraphics[height=0.157\linewidth]{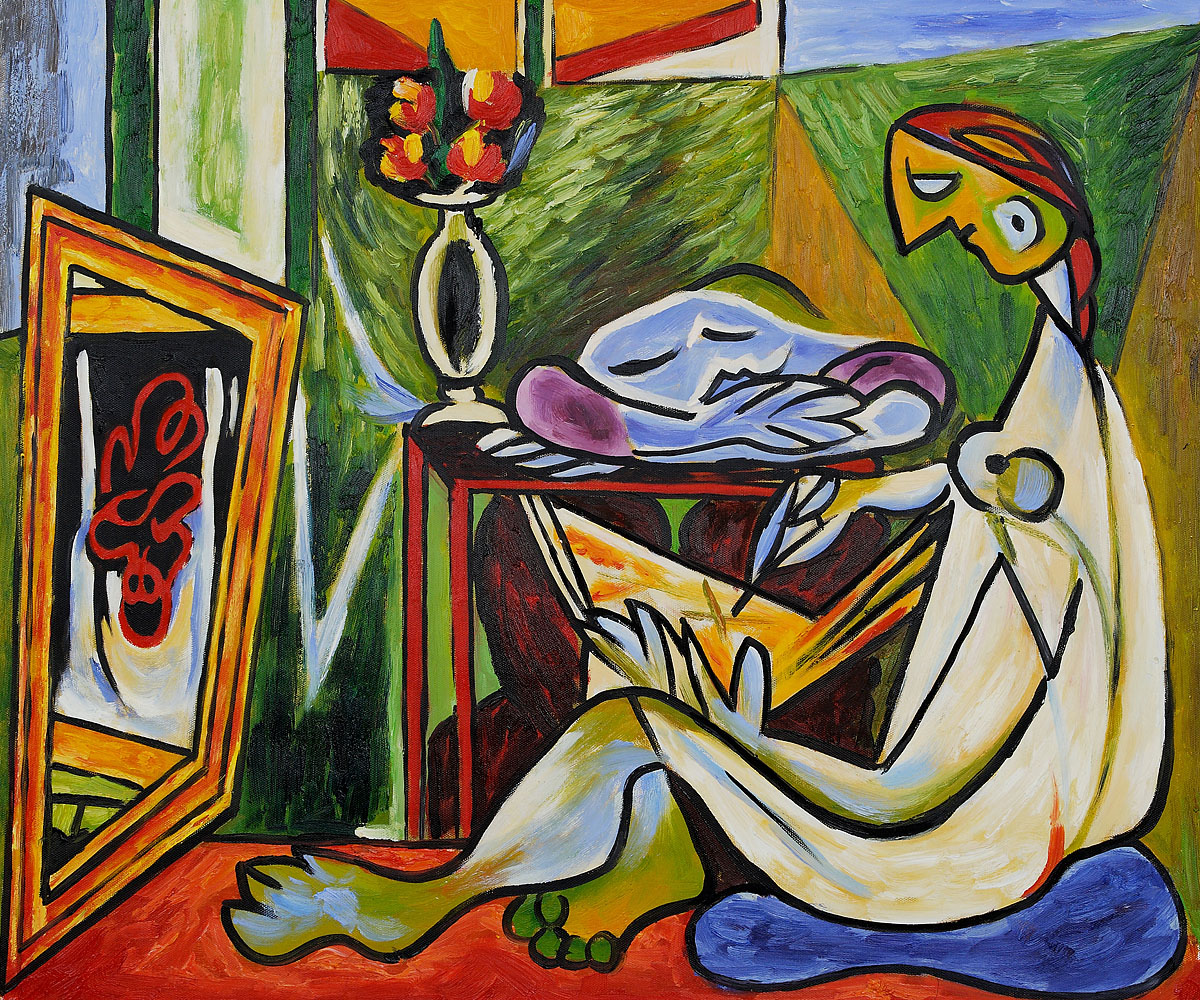}}\
		\\
		\rotatebox{90}{\makebox[20mm][c]{\small{\textsc{AdaFM}}}}
		\subfloat
		{\includegraphics[width=0.152\linewidth]{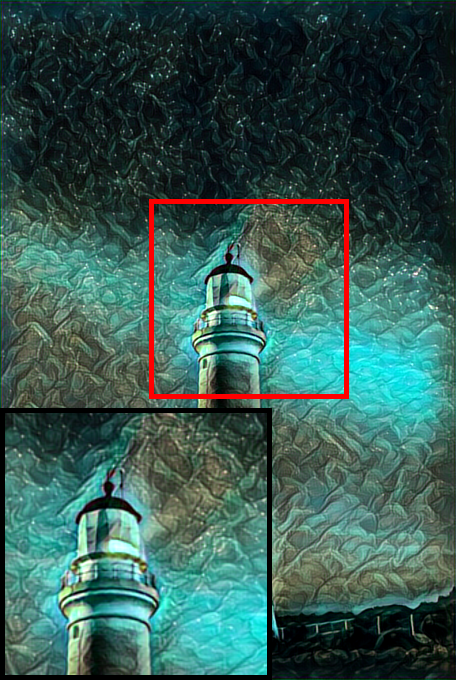}}\
		\hfill
		\subfloat
		{\includegraphics[width=0.152\linewidth]{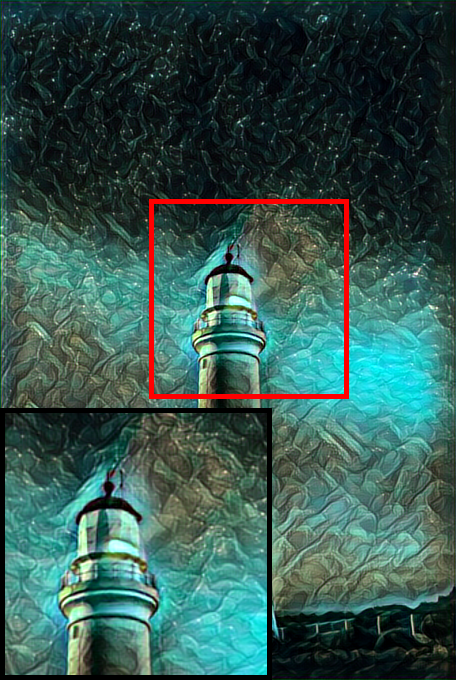}}\
		\hfill
		\subfloat
		{\includegraphics[width=0.152\linewidth]{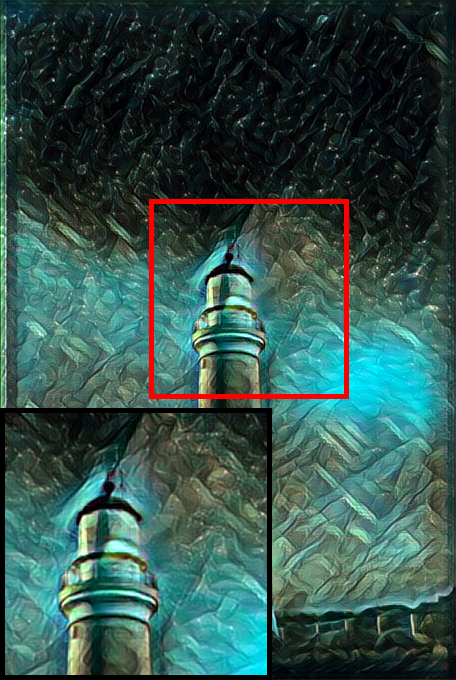}}\
		\hfill
		\subfloat
		{\includegraphics[width=0.152\linewidth]{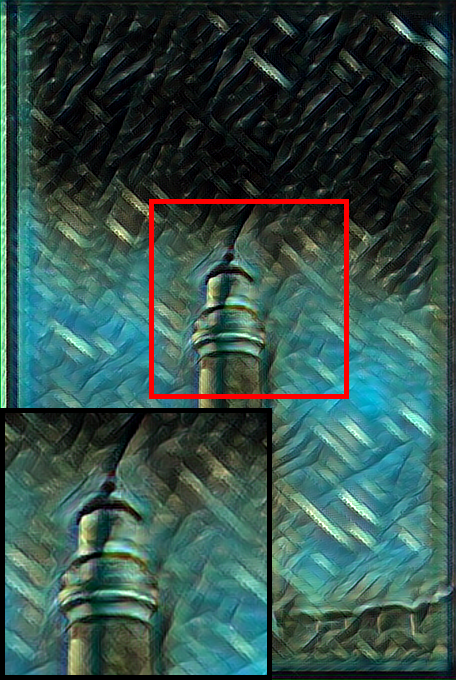}}\
		\hfill
		\subfloat
		{\includegraphics[width=0.152\linewidth]{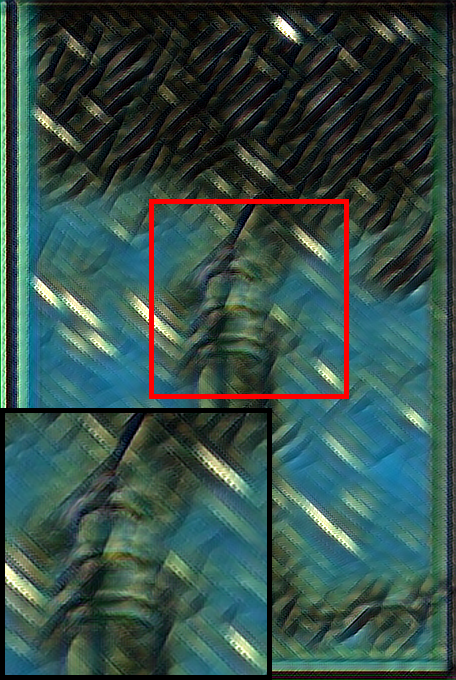}}\
		\hfill
		\subfloat
		{\includegraphics[width=0.152\linewidth]{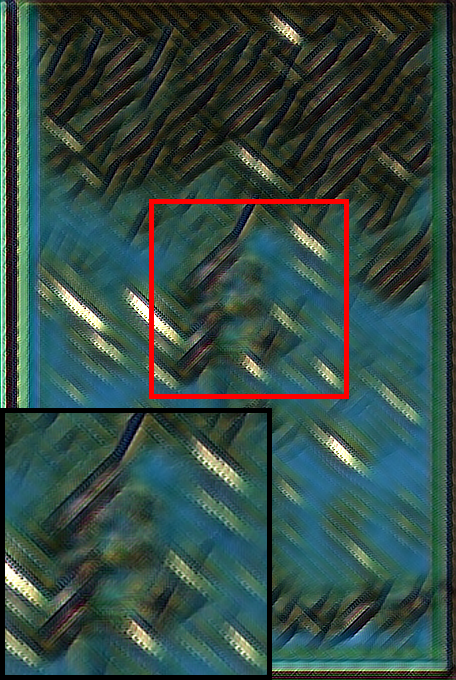}}\
		\\
		\rotatebox{90}{\makebox[20mm][c]{\small{\textsc{Dynamic-Net}}}}
		\subfloat
		{\includegraphics[width=0.152\linewidth]{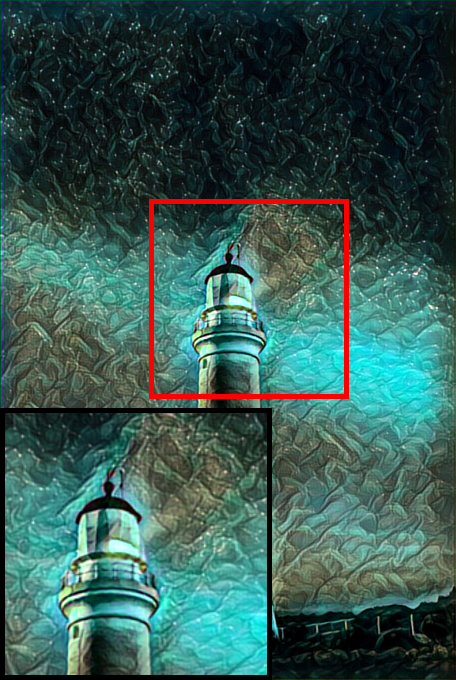}}\
		\hfill
		\subfloat
		{\includegraphics[width=0.152\linewidth]{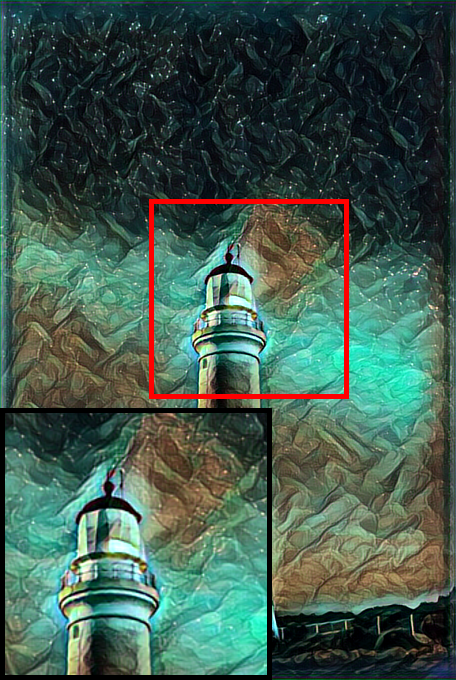}}\
		\hfill
		\subfloat
		{\includegraphics[width=0.152\linewidth]{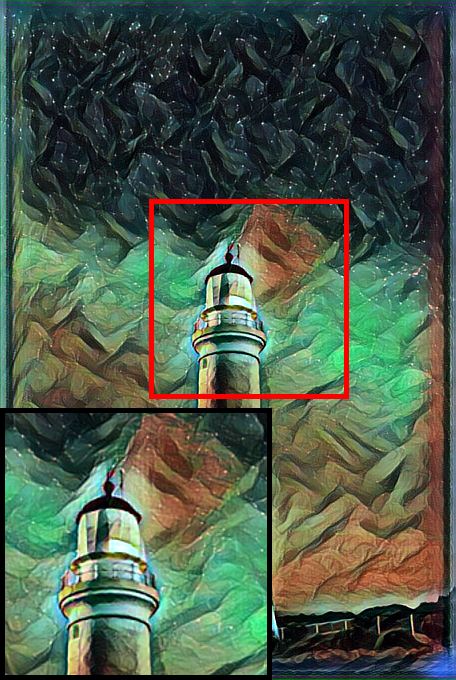}}\
		\hfill
		\subfloat
		{\includegraphics[width=0.152\linewidth]{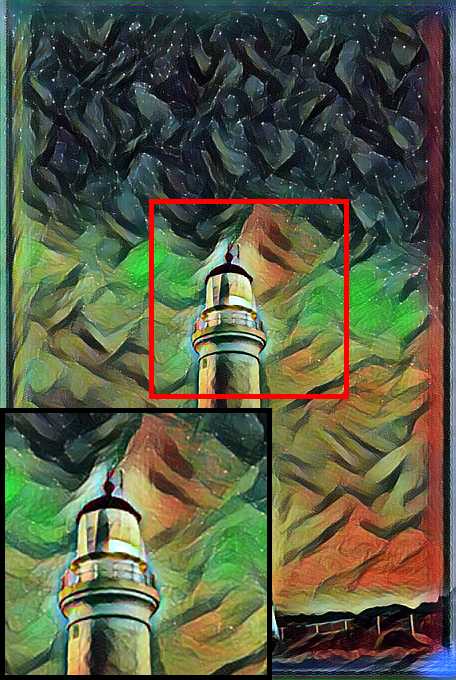}}\
		\hfill
		\subfloat
		{\includegraphics[width=0.152\linewidth]{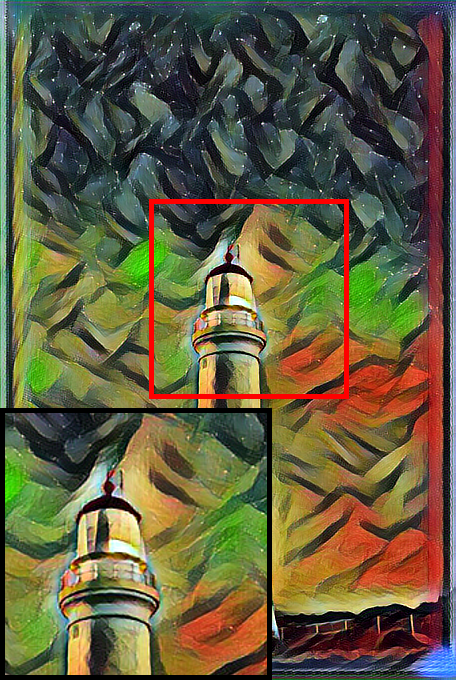}}\
		\hfill
		\subfloat
		{\includegraphics[width=0.152\linewidth]{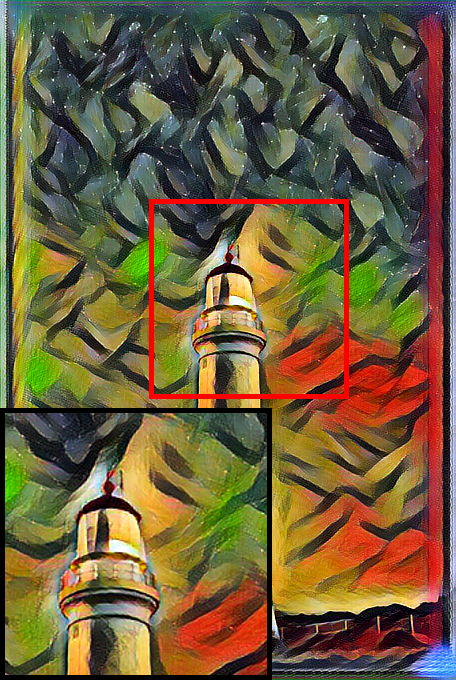}}\
		\\
		\rotatebox{90}{\makebox[20mm][c]{\small{\textsc{FTN}}}}
		\subfloat[$\alpha=0.0$]
		{\includegraphics[width=0.152\linewidth]{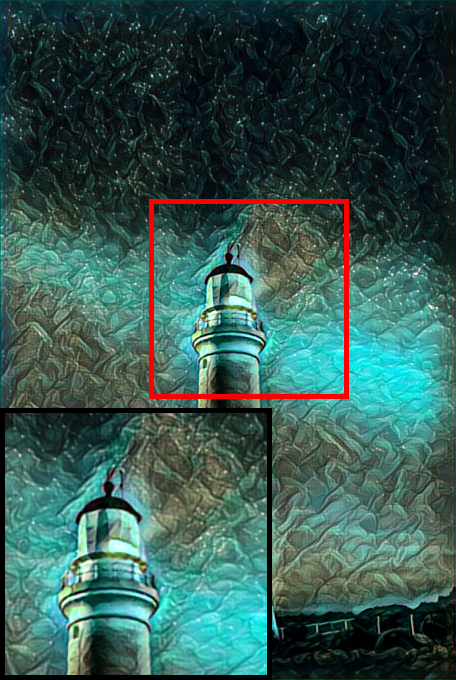}}\
		\hfill
		\subfloat[$\alpha=0.2$]
		{\includegraphics[width=0.152\linewidth]{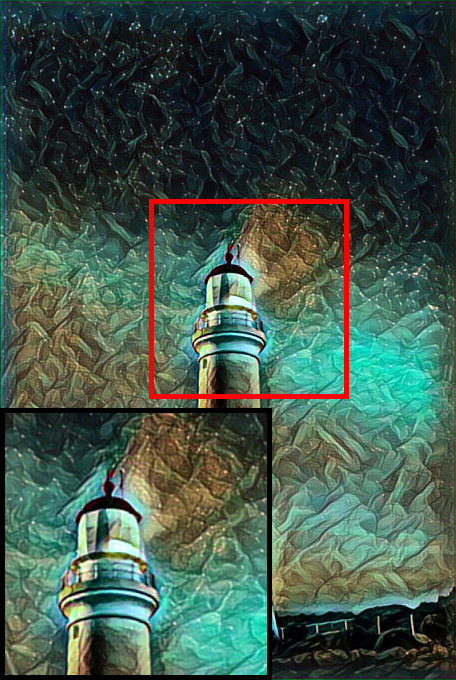}}\
		\hfill		
		\subfloat[$\alpha=0.4$]
		{\includegraphics[width=0.152\linewidth]{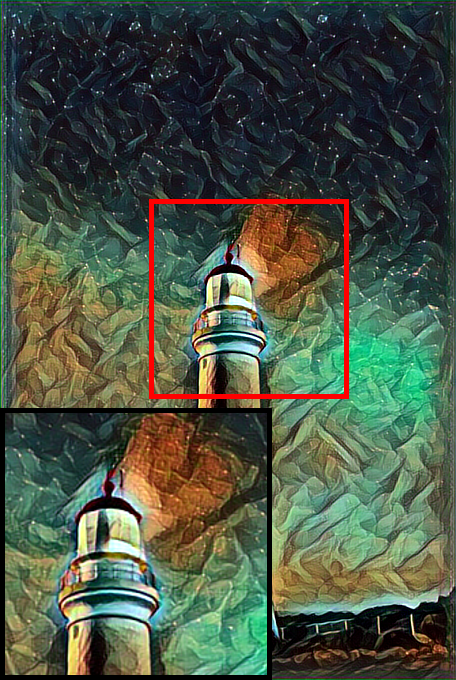}}\
		\hfill
		\subfloat[$\alpha=0.6$]
		{\includegraphics[width=0.152\linewidth]{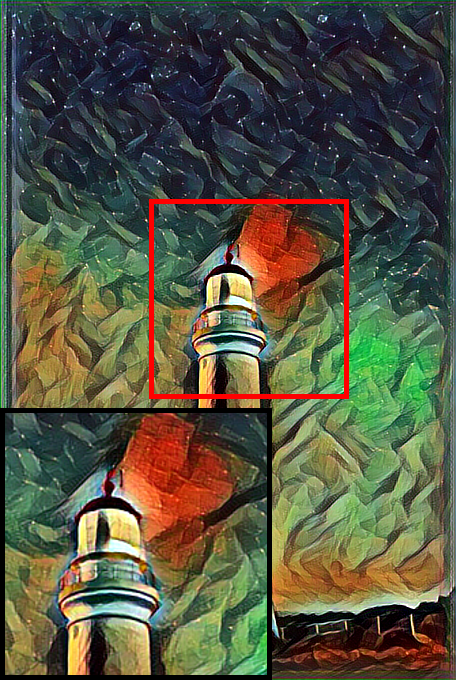}}\
		\hfill
		\subfloat[$\alpha=0.8$]
		{\includegraphics[width=0.152\linewidth]{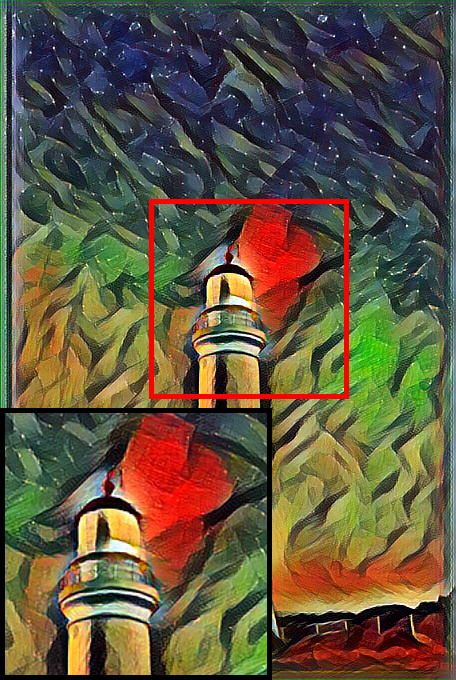}}\
		\hfill
		\subfloat[$\alpha=1.0$]
		{\includegraphics[width=0.152\linewidth]{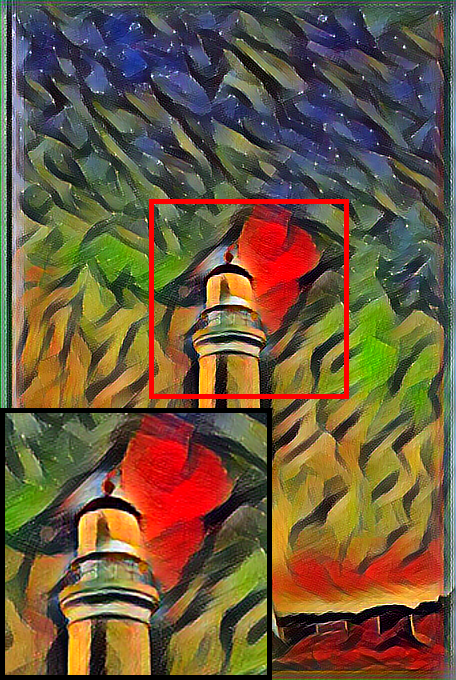}}\
		\caption{\small{\textbf{Visual comparison of controllable style transfer results between the two styles.}}}
		\label{fig:style_visual}
	\end{center}
\end{figure}

\subsubsection{Denoising \& DeJPEG.}
We use two baseline networks that were proposed in \cite{adafm2019} (AdaFM-Net) and \cite{cfsnet2019} (CFSNet-10). 
We use DIV2K \cite{DIV2k} as the training set and test on the CBSD68 \cite{cbsd68} dataset for denoising and LIVE1 \cite{live1} for deJPEG. We fine-tune the main network from the weaker noise (standard deviation 20 for denoising and quality factor 40 for deJPEG). The maximum PSNR is obtained via grid search of $\alpha$.

\subsubsection{PD-Controllable Super-resolution.}
In image super-resolution, as reported in \cite{pdt2018}, there is a trade-off between fidelity and naturalness. A comparison between algorithms should consider this trade-off by plotting perception (fidelity)-distortion (naturalness) (PD) curves. An algorithm that is closer to the origin in the PD-plane than the others implies it has better performance. Drawing this PD-curve is possible by changing the weights between loss terms. As in \cite{esrgan2018}, we first-train the network using $L_{1}$ loss and second-train it using a combined loss of $L_{1}$, Perceptual \cite{perceptual2016}, and GAN \cite{gan2014} losses. We evaluate using two baseline networks that were proposed for \cite{cfsnet2019} (CFSNet-30) and \cite{esrgan2018} (ESRGAN). We use DIV2K as the training set and PIRM \cite{pirm2018} as the test set. PSNR and SSIM \cite{ssim} are used as distortion metrics, and NIQE \cite{niqe} and the Perceptual Index \cite{pirm2018} are used as perception metrics. 

\subsubsection{Style Transfer.}
In style transfer, we use Transform-Net which was proposed in \cite{perceptual2016} with instance normalization \cite{instance_norm}. We follow the settings of Dynamic-Net \cite{dynamicnet2019}. The COCO 2014 train dataset \cite{cocodataset} is used for training. From the main network, Dynamic-Net inserts three tuning branches into pre-defined layers, while FTNs are inserted in every convolution layer. This means that FTNs have more opportunities to control in a layer-wise manner (Fig. 1 of the supplementary material of Dynamic-Net \cite{dynamicnet2019}).

\begin{table}[!t]
	\centering
	\caption{{\textbf{Gaussian Denoising Results.} Average PSNR (dB) on CBSD68 test dataset. Unseen noise levels are denoted with *. The best results are \textbf{bold-faced} and the second best results are \underline{underlined}.}}
	\resizebox{1.0\linewidth}{!}{
		\begin{tabular}{c|cccc|cccc}
			\toprule
			\multicolumn{1}{c|}{} & \multicolumn{4}{c|}{\large{CFSNet-10 (10 Blocks)}} & \multicolumn{4}{c}{\large{AdaFM-Net (16 Blocks)}}  \\ 
			\midrule 
			Noise Level & 20  &  40*   & 60*  & 80	& 20   & 40*   & 60*  & 80  \\
			\midrule
			From Scratch & 32.42 & 28.98 & 27.13 & 25.90 & 32.44 & 28.90 & 26.92 & 25.60 \\
			\midrule
			DNI & \textbf{32.42} & \textbf{28.87} & \textbf{27.01} & \textbf{25.96} & \textbf{32.44} & 28.20 & \underline{26.98} & 25.97 \\
			AdaFM & \textbf{32.42}& 28.48 & 26.75 & 25.84 & \textbf{32.44} & 28.17 & 26.77 & 25.96  \\
			CFSNet & \textbf{32.42} & 28.65 & \underline{26.95} & \underline{25.93} & \textbf{32.44} & 28.41 & 26.87 & \underline{26.00}  \\
			\midrule
			\textbf{FTN-gc16} & \textbf{32.42} & \underline{28.77} & \textbf{27.01} & 25.89 & \textbf{32.44} & \textbf{28.78} & \textbf{27.05} & 25.98 \\
			\textbf{FTN-gc4} & \textbf{32.42} & 28.65 & 26.90 & 25.90 & \textbf{32.44} & \underline{28.64} & 26.95 & \underline{26.00} \\
			\textbf{FTN} & \textbf{32.42} & 28.45 & 26.86 & \underline{25.93} & \textbf{32.44} & 28.48 & 26.89 & \textbf{26.03}  \\ 
			\hline
	\end{tabular}}
	\label{tb:dn_result}
\end{table}

\begin{figure}[!t]
	\begin{center}
		\captionsetup[subfigure]{labelformat=empty}
		\rotatebox{90}{\makebox[22mm][c]{{\textsc{CFSNet-30}}}}
		\subfloat
		{\includegraphics[width=0.450\linewidth]{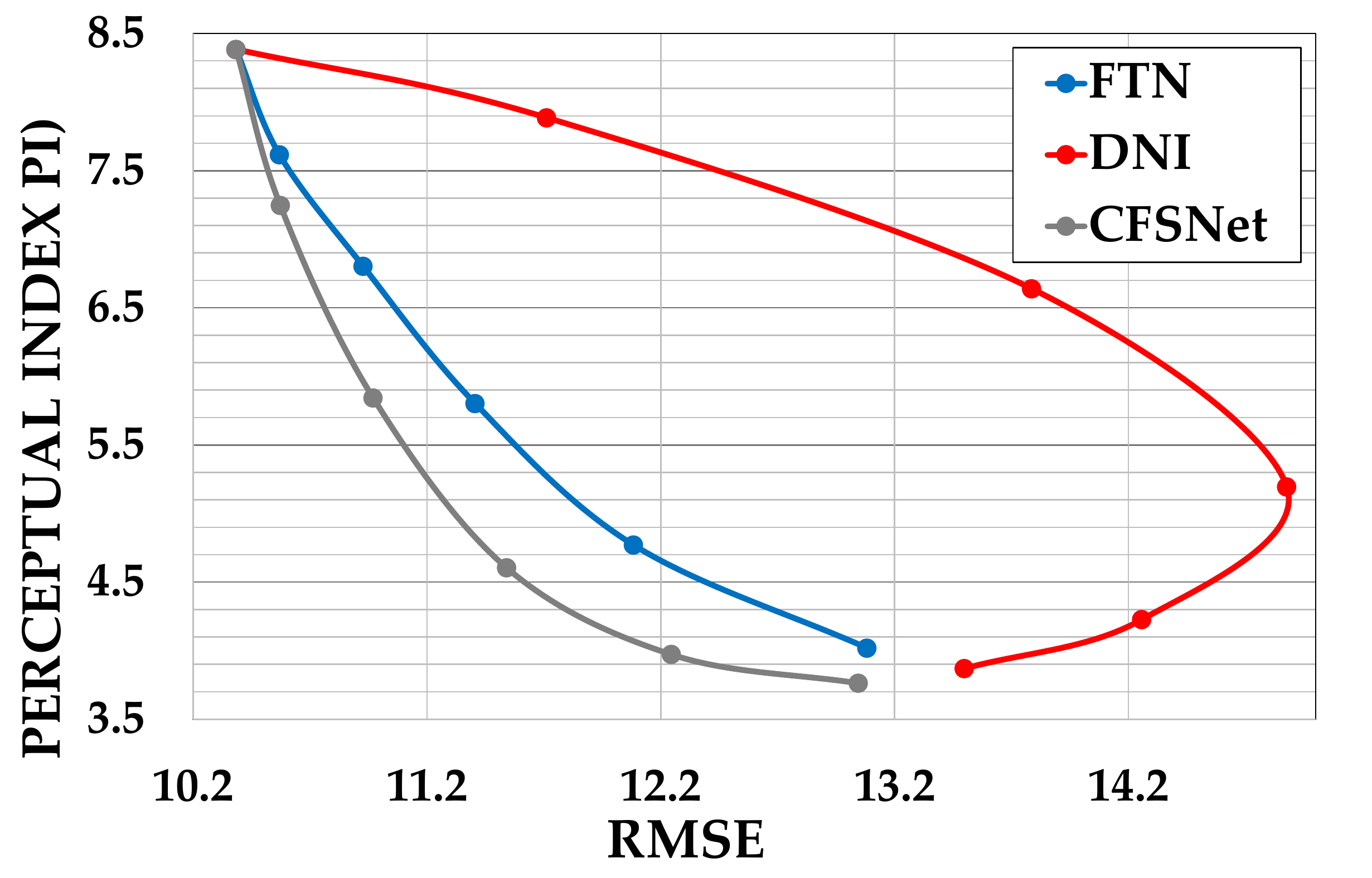}}\
		\subfloat
		{\includegraphics[width=0.450\linewidth]{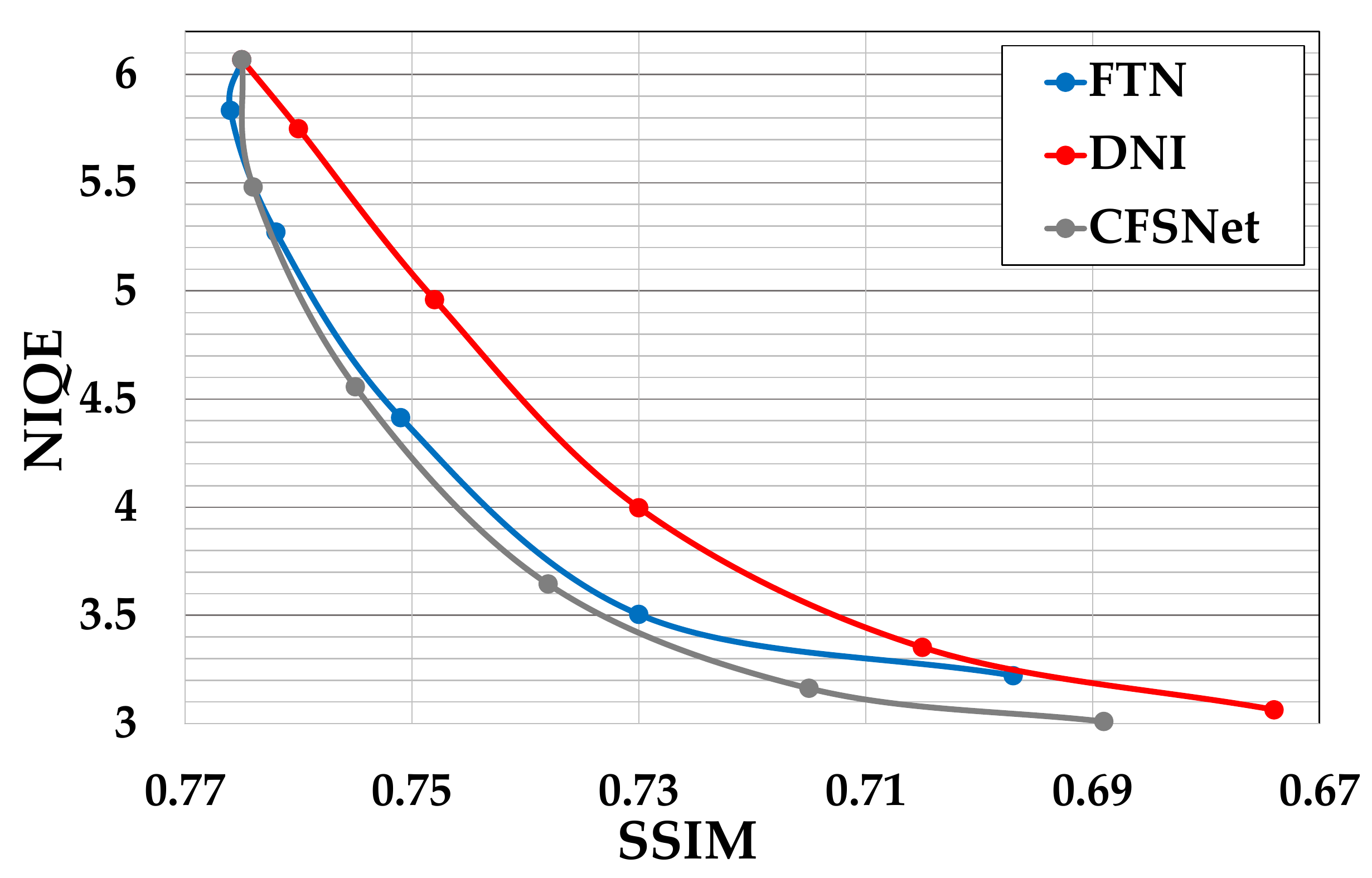}}\
		\\
		\rotatebox{90}{\makebox[22mm][c]{{\textsc{ESRGAN}}}}
		\subfloat[{RMSE-PI}]
		{\includegraphics[width=0.450\linewidth]{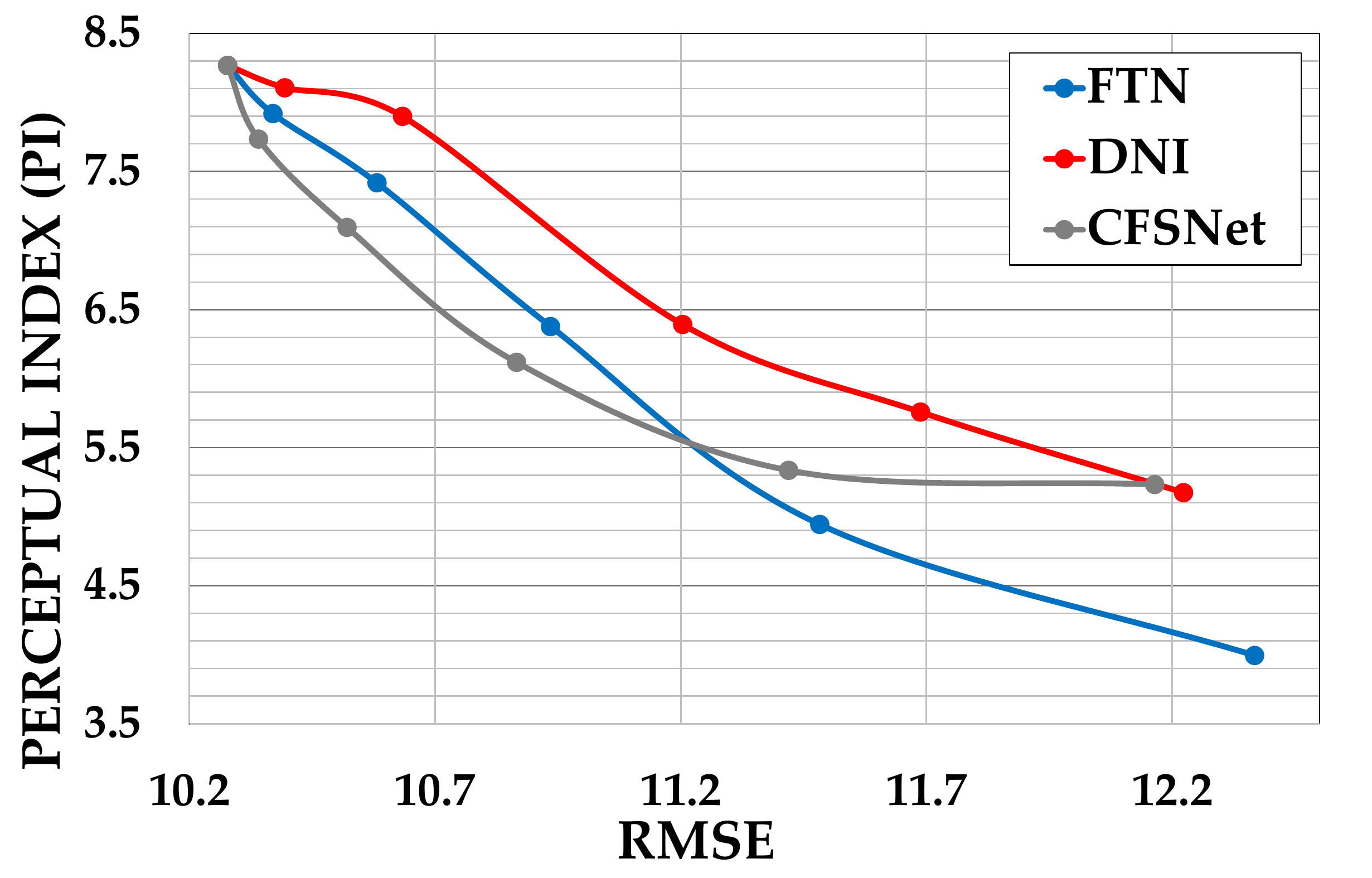}}\
		\subfloat[{SSIM-NIQE}]
		{\includegraphics[width=0.450\linewidth]{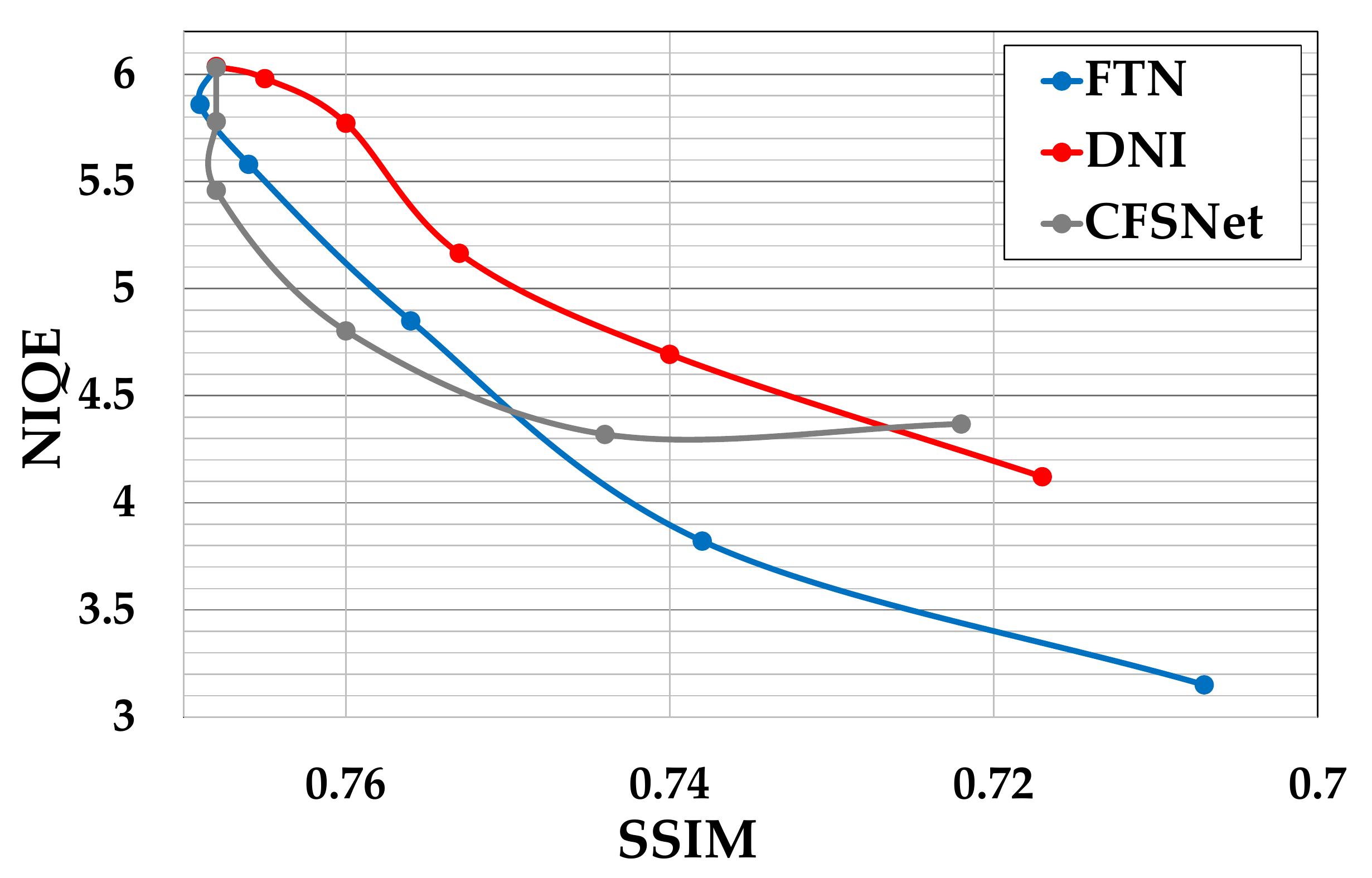}}\
		\caption{\small{\textbf{Results of PD-controllable image super-resolution ($\times$4).} Combined, various adaptation results, and results images are described in supplementary material.}}
		\label{fig:sr_results}
	\end{center}
\end{figure}

\subsection{Ablation Study}
\label{ablation}

We perform an ablation study on AdaFM-Net for image denoising to compare different structures of FTNs in Table \ref{tb:dn_ablation}. 
We define two additional versions of FTN: FTN-\textit{deeper} and FTN-\textit{spatial}. 
FTN-\textit{deeper} is a three-layer version of the FTN whose intermediate results are worse than others because excessive modification of the filters affects the interpolation results. FTN-\textit{spatial} is a depth-wise convolution version whose performance is inferior to other channel-wise convolutions. 

\subsection{Adaptation and Interpolation Performance}
\label{adaptation}

Adaptation performance refers to the performance on the tuned second level compared to a network trained only for the level. The interpolation performance refers to the performance on the unseen intermediate interpolated levels. The degree of performance depends on the evaluation metrics of each task. 

First, Fig.~\ref{fig:style_visual} depicts the result of the style transfer task, which requires a large transition of the model parameters as the style changes. 
According to Fig.~\ref{fig:style_visual}, AdaFM has difficulty adapting from style A to the style B. 
These results show that linear adaptation has limitations in reaching the hard second level. 
In Dynamic-Net, it cannot deliver the second style smoothly because it only changes three pre-defined layers while FTN changes all convolutional filters. 
More results for stylization are described in the supplementary material.

\begin{figure}[!t]
	\begin{center}
		\includegraphics[width=1.0\linewidth]{"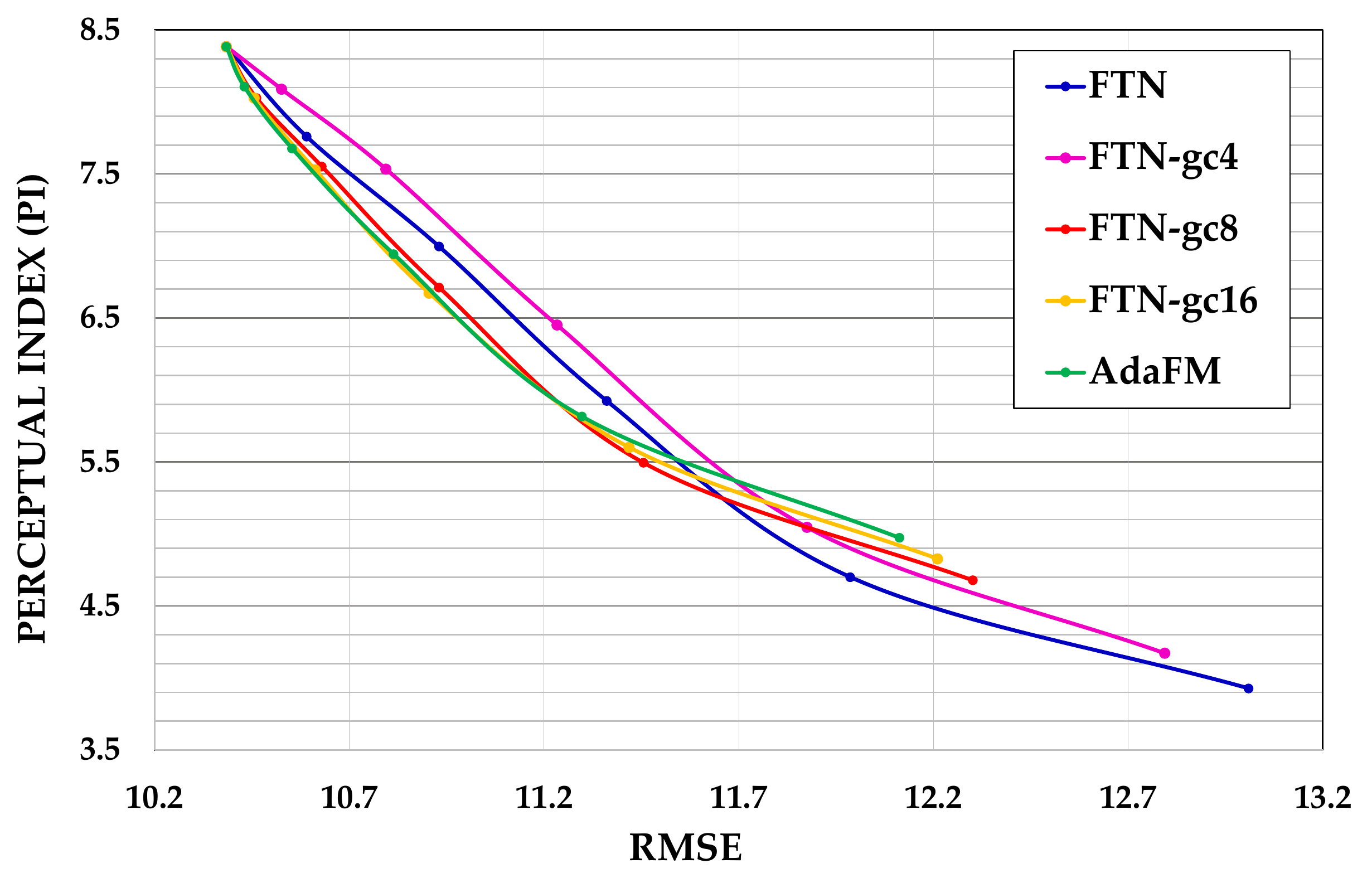}
	\end{center}
	\caption{\small{\textbf{Ablation study for group convolutions in PD-control}.}}
	\label{fig:sr_abl}	
\end{figure}

Subsequently, Table~\ref{tb:dn_result} shows the adaptation/interpolation performances on the denoising task (Results images and results on deJPEG task are reported in the supplementary material.). 
The table shows that there is not notable difference between performances over the compared methods, including AdaFM which uses linear adaptation. 
This is because denoising tasks require their model parameters to be changed less as the level changes compared to other tasks (\textit{e.g.} style transfer). 
FTN-gc4, 16 indicate the group convolution version of FTN with 4 and 16 groups, respectively. They show better interpolation performances at the expense of adaptation performance. 
Specifically, in AdaFM-Net, FTN-gc4 and FTN-gc16 outperforms the other frameworks. In CFSNet-10, DNI outperforms other frameworks but the margin is not large. In CFSNet-10, the network is shallower (10 ResBlocks) than AdaFM-Net (16 ResBlocks), which means that the parameter space can be easily linear. This can increase the filter similarity of simple fine-tuning (DNI). Compared to AdaFM and CFSNet, the performance of FTNs is superior.

\begin{table}[!b]
	\centering
	\caption{\small{\textbf{Filter analysis for regularization.} Distance and Similarity between the two levels. We measure Mean Average Error (MAE) for linear filter interpolation and filter-wise normalized cosine similarity. The task is PD-controllable super-resolution and baseline network is \textbf{CFSNet-30}}}
	\resizebox{0.75\linewidth}{!}{
		\begin{tabular}{cc|ccc}
			\toprule
			&   & FTN (G=16)  & FTN  & Fine-tuning  \\ \midrule	
			& MAE  & \textbf{0.0082}  & 0.0118  & 0.0139  \\
			& Cos Sim.  & \textbf{0.9443} & 0.8937 & 0.8666  \\ \bottomrule	
	\end{tabular}}
	\label{tb:sr_similarity}
\end{table}

\begin{figure*}[!pt]
	\begin{center}
		\captionsetup[subfigure]{labelformat=empty}
		\subfloat[(a) FTN-gc16]
		{\includegraphics[width=0.330\linewidth]{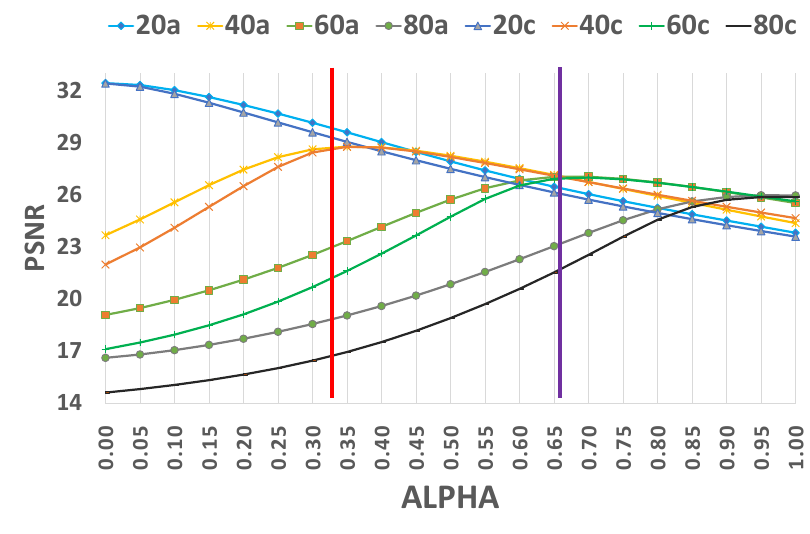}}\
		\subfloat[(b) CFSNet]
		{\includegraphics[width=0.330\linewidth]{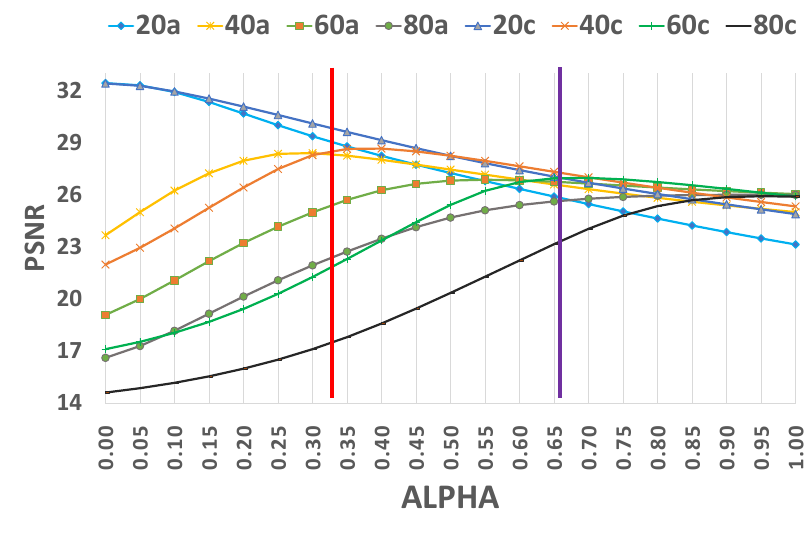}}\
		\subfloat[(c) DNI]
		{\includegraphics[width=0.330\linewidth]{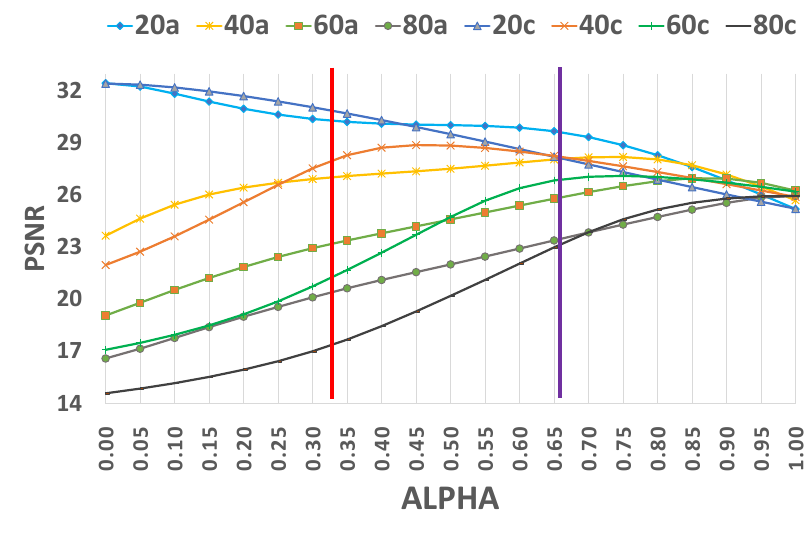}}\
		\caption{\small{\textbf{Smoothness analysis for denoising. ($\sigma=20$ to $\sigma=80$)} We plot {\textbf{$\sigma=40$}} (red) and {\textbf{$\sigma=60$}} (purple) lines as linearly optimal interpolation points. For each curve, number indicates input quality factor, \textbf{$a$} denotes AdaFM-Net network and \textbf{$c$} denotes CFSNet-10 network. For example, \textit{40a} indicates $\sigma=40$ results on AdaFM-Net. Our FTN-gc16 results show that the choice of $\alpha$ is closest to the lines}}
		\label{fig:graph_dn_network}
	\end{center}
\end{figure*}

However, compared to the denoising task, the DNI shows a different pattern in the PD-control (Fig.~\ref{fig:sr_results}). 
Although DNI performs well for both end levels, it shows significantly unstable and low performance for the intermediate levels. This is because the fine-tuning process of the DNI simply updates the parameters, without considering the initial state. Therefore, the relation between the parameters for the two levels becomes weaker compared to simple denoising tasks, and the interpolated parameters start not to behave as intended. A detailed analysis will be described in the following section.

\begin{figure}[!t]
	\begin{center}
		\captionsetup[subfigure]{labelformat=empty}
		\hfill
		\rotatebox{90}{\makebox[28mm][c]{{\textsc{HR}}}}
		\subfloat
		{\includegraphics[width=0.450\linewidth]{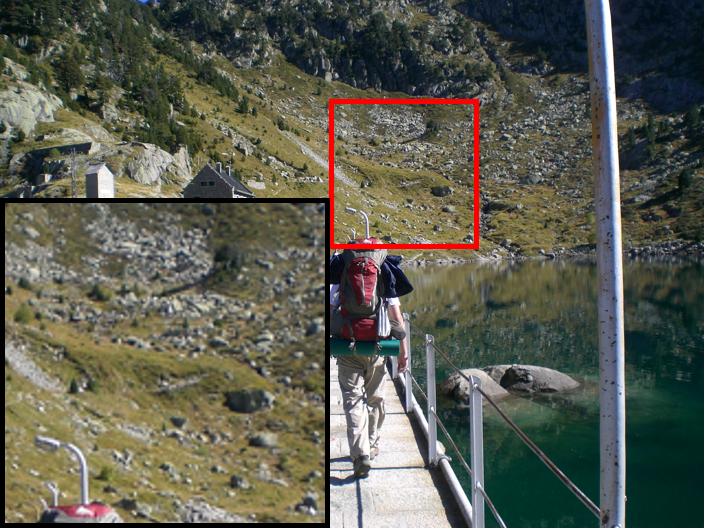}}\
		\\
		\rotatebox{90}{\makebox[28mm][c]{{\textsc{FTN}}}}
		\subfloat
		{\includegraphics[width=0.450\linewidth]{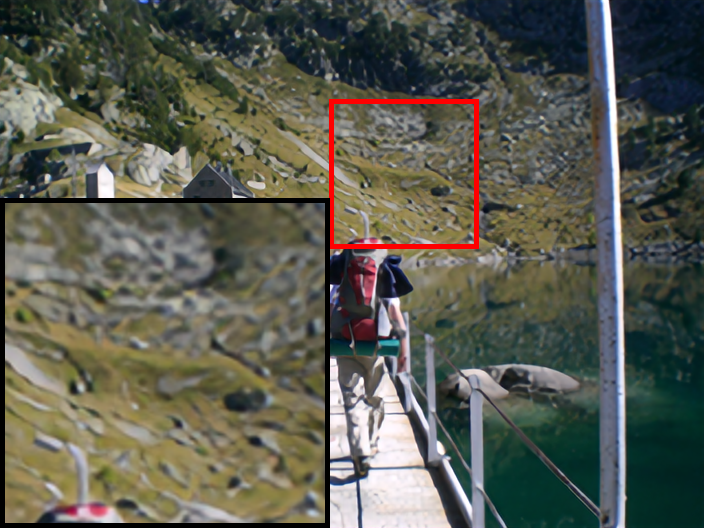}}\
		\hfill
		\subfloat
		{\includegraphics[width=0.450\linewidth]{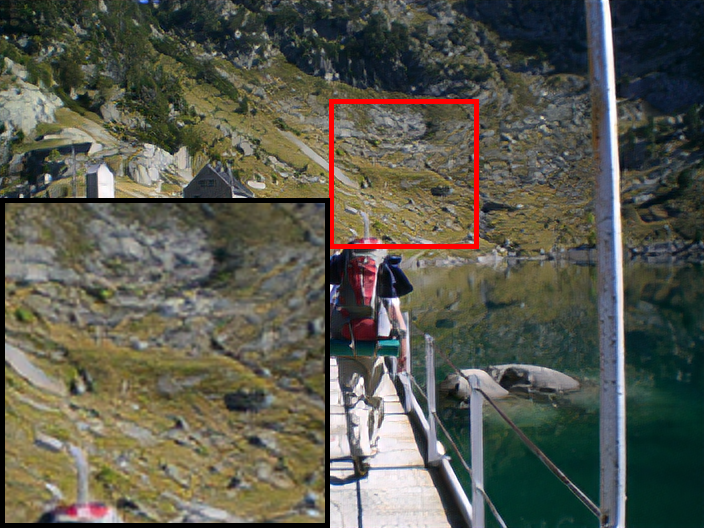}}\
		\\
		\rotatebox{90}{\makebox[28mm][c]{{\textsc{DNI}}}}
		\subfloat[(a) $\alpha=0.6$]
		{\includegraphics[width=0.450\linewidth]{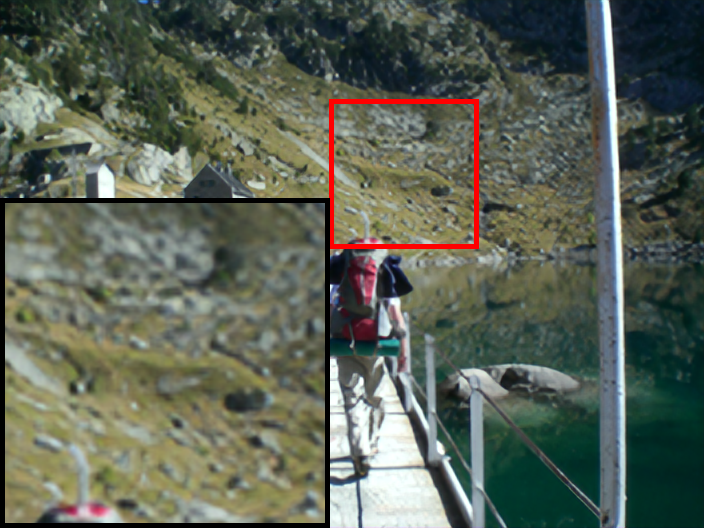}}\
		\hfill
		\subfloat[(b) $\alpha=1.0$]
		{\includegraphics[width=0.450\linewidth]{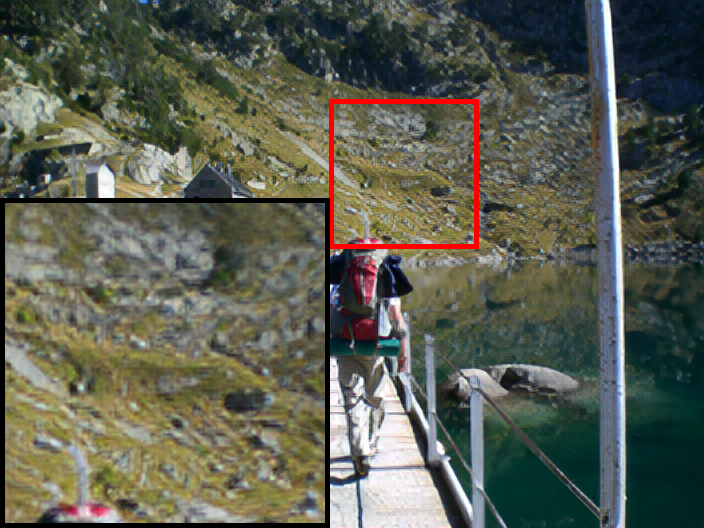}}\
		\caption{\textbf{Perception-distortion controllable super-resolution results.}}
		\label{fig:sr_results2}
	\end{center}
\end{figure}

\subsection{Smoothness}
\label{interpolation}

In this section, we empirically analyze the definition of \textit{smoothness} (in the introduction) and prove our filter similarity assumption that keeping the filters similar to the first level ($\textbf{f}^{(1)}_{i}$) improves smoothness performance (in the proposed approach).

\subsubsection{Filter Similarity.}
FTNs are designed to keep the original filters when they are tuned non-linearly to the second level. To verify this, we measure filter similarity in terms of absolute distance and cosine similarity for the super-resolution task. Table \ref{tb:sr_similarity} shows the filter distance with the filters of the main network when they are fine-tuned (DNI) or passed through FTNs. The results show that filter-conditioned tuning is effective in preventing significant filter change, and using group convolution further restricts filter changes.

\subsubsection{Group Convolution.}
Group convolution in the FTNs restricts filter changes and this restriction guarantees smoothness. 
Fig.~\ref{fig:sr_abl} verifies this from the results of FTNs by changing the number of groups and AdaFM (linear version). The curves prove that the large filter similarity exhibits better interpolation performance at the expense of the second-level adaptation performance. 

\subsubsection{Color Distortion.}
In Fig.~\ref{fig:sr_results}, DNI indicates unstable intermediate results in some metrics. 
From the intermediate images in Fig.~\ref{fig:sr_results2} for DNI, a slight color difference can cause significant pixel-error (RMSE), but a similar value in the SSIM metric. In contrast, FTN has no such color distortions. The full results are described in the supplementary material. 

\subsubsection{Interpretability.} 
For practical use, it will be essential for the users to know which value of $\alpha$ corresponds to which level. For example, in the denoising task, suppose that we train a network to work between the levels $\sigma=20$ and $\sigma=80$. When we set $\alpha=0.5$, it is reasonable that the network will perform best for the level $\sigma=50$, which is the middle point of the interval. In other words, $\alpha$ must be linear along with the level. Fig.~\ref{fig:graph_dn_network} shows the result of the denoising task over various noise level $\sigma$ of the test set and the parameter $\alpha$. According to the figure, the maximum performance of our FTN for $\sigma=40$ and $\sigma=60$ best matches the vertical lines of $\alpha=0.33$ and $\alpha=0.66$, compared to the other methods.

\subsubsection{Over-smoothing Artifacts.} 
\begin{figure}[!t]
	\begin{center}
		\captionsetup[subfigure]{labelformat=empty}
		\rotatebox{90}{\makebox[20mm][c]{\small{\textsc{Input}}}}
		\subfloat
		{\includegraphics[width=0.158\linewidth]{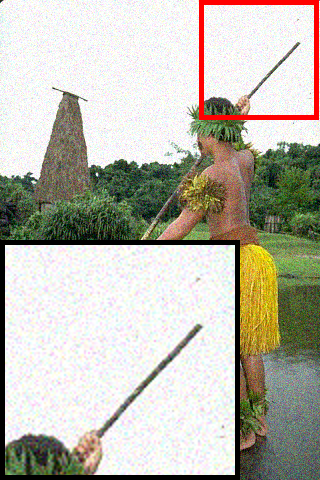}}\
		\hfill
		\rotatebox{90}{\makebox[20mm][c]{\small{\textsc{Clean}}}}
		\subfloat
		{\includegraphics[width=0.158\linewidth]{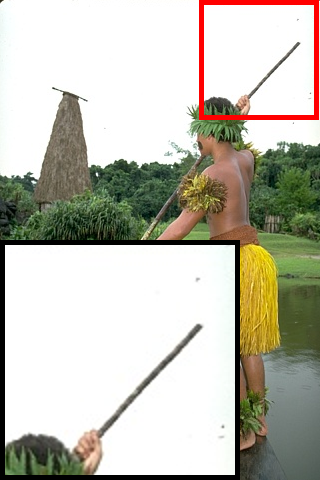}}\
		\\
		\rotatebox{90}{\makebox[13mm][c]{\small{\textsc{FTN-gc16}}}}
		\subfloat
		{\includegraphics[width=0.152\linewidth]{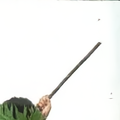}}\
		\hfill
		\subfloat
		{\includegraphics[width=0.152\linewidth]{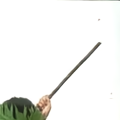}}\
		\hfill		
		\subfloat
		{\includegraphics[width=0.152\linewidth]{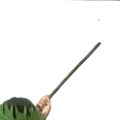}}\
		\hfill
		\subfloat
		{\includegraphics[width=0.152\linewidth]{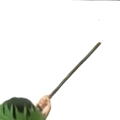}}\
		\hfill
		\subfloat
		{\includegraphics[width=0.152\linewidth]{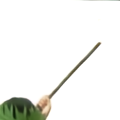}}\
		\hfill
		\subfloat
		{\includegraphics[width=0.152\linewidth]{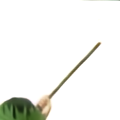}}\
		\\
		\rotatebox{90}{\makebox[13mm][c]{\small{\textsc{DNI}}}}
		\subfloat
		{\includegraphics[width=0.152\linewidth]{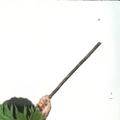}}\
		\hfill
		\subfloat
		{\includegraphics[width=0.152\linewidth]{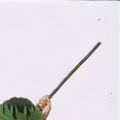}}\
		\hfill		
		\subfloat
		{\includegraphics[width=0.152\linewidth]{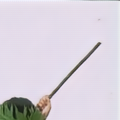}}\
		\hfill
		\subfloat
		{\includegraphics[width=0.152\linewidth]{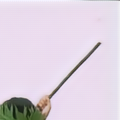}}\
		\hfill
		\subfloat
		{\includegraphics[width=0.152\linewidth]{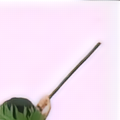}}\
		\hfill
		\subfloat
		{\includegraphics[width=0.152\linewidth]{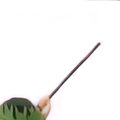}}\
		\\
		\rotatebox{90}{\makebox[13mm][c]{\small{\textsc{CFSNet}}}}
		\subfloat[$\alpha=0.0$]
		{\includegraphics[width=0.152\linewidth]{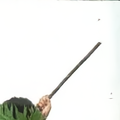}}\
		\hfill
		\subfloat[$\alpha=0.2$]
		{\includegraphics[width=0.152\linewidth]{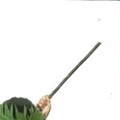}}\
		\hfill		
		\subfloat[$\alpha=0.4$]
		{\includegraphics[width=0.152\linewidth]{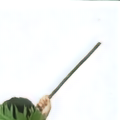}}\
		\hfill
		\subfloat[$\alpha=0.6$]
		{\includegraphics[width=0.152\linewidth]{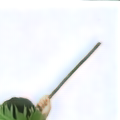}}\
		\hfill
		\subfloat[$\alpha=0.8$]
		{\includegraphics[width=0.152\linewidth]{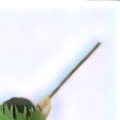}}\
		\hfill
		\subfloat[$\alpha=1.0$]
		{\includegraphics[width=0.152\linewidth]{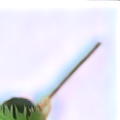}}\
		\caption{\small{\textbf{Denoising results on weak noise level ($\sigma=20$). When user controls $\alpha$ to a large value, over-smoothing artifacts arise}}}
		\label{fig:dn_results}
	\end{center}
\end{figure}

In real-world applications, because the user may not know the degradation level, the user hopes to control the \textit{strength} of the denoising. We describe our visual denoising result in an extreme case as presented in Fig.~\ref{fig:dn_results}. In Fig.~\ref{fig:dn_results}, the input noise level is 20, which means that the optimal results come from $\alpha=0$ in all frameworks. $\alpha=1$, which is optimal for noise level 80, can over-smoothen the image. When $\alpha$ increases, the DNI and CFSNet results reveal some color artifacts in the background, while the FTN-gc16 results are much cleaner, which is significant for the real-world feedback-based systems. Because CFSNet exploits a dual network structure, each network cannot consider the other level in the test phase. In contrast, in FTNs, two filters for both sides of the FTNs are highly correlated.

\subsection{Efficiency}
\subsubsection{Complexity}
If any tuning layer with convolutions on a feature map is added, additional computations (MACs) are $H \times W \times K_{H} \times K_{W} \times C_{in} \times C_{out}$ where $H$, $W$, $K_{H}$, $K_{W}$, $C_{in}$ and $C_{out}$ are the height and width of the feature map (e.g., image size), height and width of the filter, and the number of input channels and output channels, respectively. Dominant computations arise from $H$ and $W$. In our network, which is a data-independent module, only $K_{H} \times K_{W} \times C_{in} \times (C_{out}/Groups) \times N$ is needed for a single tuning layer, where $N$ is the depth of the FTNs. As shown in Table \ref{tbl:complexity}, FTNs have extremely reduced computational complexity and a similar or much lower number of parameters than other frameworks in various tasks and networks.

\subsubsection{Pixel-adaptive Extension.}
\begin{figure}[!t]
	\begin{center}
		\includegraphics[width=1.0\columnwidth]{"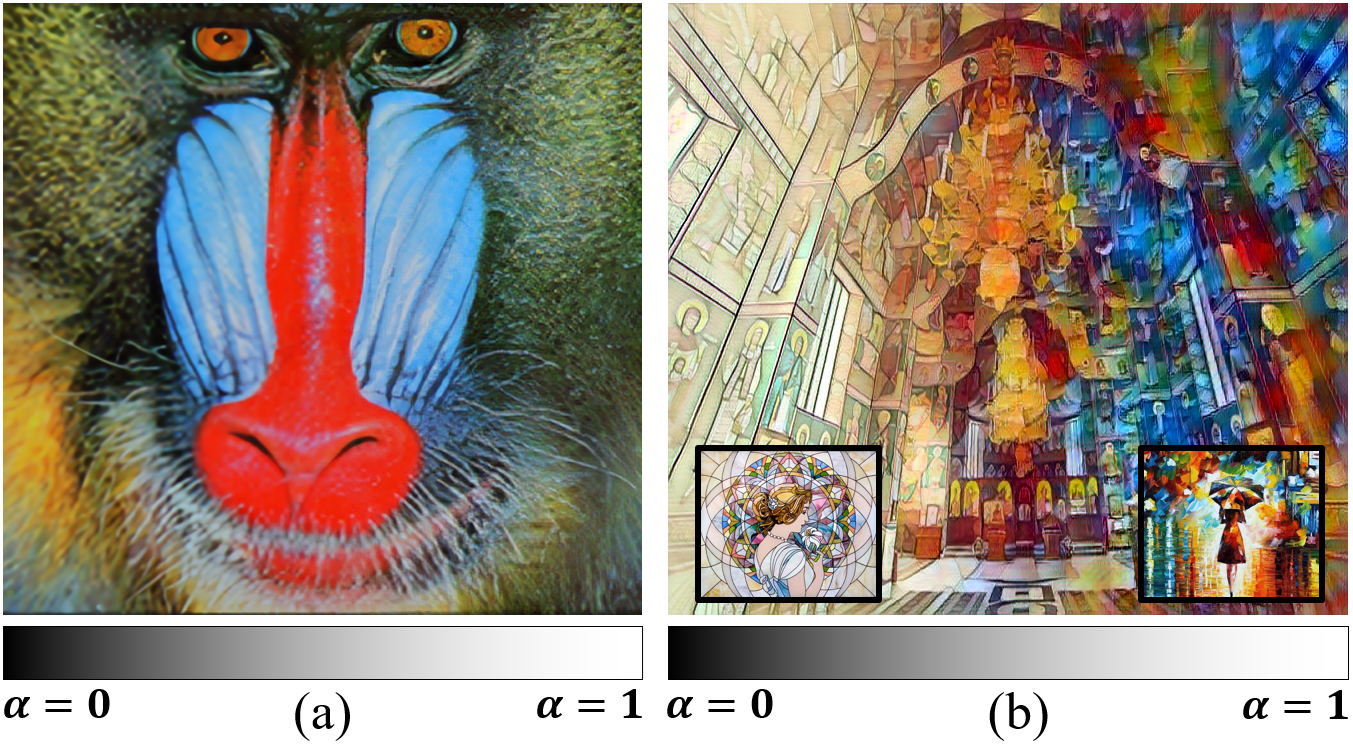}
	\end{center}
	\caption{\small{\textbf{Pixel-adaptive control results.} Zoom in for a better view}}
	\label{fig:pixel_adaptive}
\end{figure}

Considering real-world imaging applications, the user wants to control not only the global level but also locally (pixel-wise). In this case, every pixel has its own imaging levels from $\alpha=0$ to $\alpha=1$. Naive pixel-adaptive control requires filters for every level, which can cause large memory issues. For efficient inference of pixel-adaptive continuous control, we propose a simple modification from the pixel-adaptive convolution. This is described as follows.

\begin{equation}
\begin{aligned}
\textbf{Y} {} & = \textbf{X} {*}_{i,j} (\textbf{f} \times (1-{\alpha}_{i, j}) + FTN(\textbf{f}) \times {\alpha}_{i, j} ) \\
& = (\textbf{1}-{\textbf{A}}) \odot (\textbf{X} * \textbf{f})  + {\textbf{A}} \odot (\textbf{X} * FTN(\textbf{f})  )
\end{aligned}
\end{equation}

\noindent where $\textbf{f}$ is the global filter, $\textbf{*}_{i, j}$ is the pixel-adaptive convolution, ${\alpha}_{i, j}$ is the per-pixel level, and \textbf{A} is the global level map that describes the pixel-wise levels. $\odot$ denotes element-wise multiplication. This modification makes implementation much simpler because only two global convolutions and multiplications are needed for pixel-adaptive control.
Examples are shown in Fig. \ref{fig:pixel_adaptive}. We test two examples: \textit{PD-control} and \textit{style control}. In Fig. \ref{fig:pixel_adaptive} (a), from the leftmost pixels to the rightmost pixels, the PSNR decreases and the texture becomes sharper (higher perceptual quality) continuously. In Fig. \ref{fig:pixel_adaptive} (b), the pixels are smoothly stylized from one style to the other. More results with high-resolution sources can be found in the supplementary material.

\begin{table}[!t]
	\begin{center}
		\caption{\small{Overall computations, relative computations from baseline (in percentage), and number of parameters of the frameworks.}}
		\resizebox{1.0\linewidth}{!}{
			\begin{tabular}{lcccc}
				\toprule 
				Task & \multicolumn{2}{c}{Denoising} & \multicolumn{2}{c}{$\times$2 Super-Resolution} \\
				\midrule
				Network & \multicolumn{2}{c}{AdaFM-Net} & \multicolumn{2}{c}{CFSNet-30} \\
				\cmidrule{2-5}          & GFLOPs  & Params(M)  & GFLOPs  & Params(M) \\
				\midrule
				Baseline & 25.11  & 1.41 & 155.96  & 2.37  \\
				+ CFSNet & 46.96 (87.02\%)  & 3.06 &  311.36 (99.64\%)  & 4.93    \\
				+ AdaFM  & 26.01 (3.58\%)  & \textbf{1.46}  & 162.50 (4.20\%)  & \textbf{2.47}  \\
				+ \textbf{FTN} & \textbf{25.36 (0.10\%)} & 1.83 & \textbf{156.34 (0.02\%)} & 3.01 \\
				+ \textbf{FTN(G=16)} & \textbf{25.13 (0.01\%)} & \textbf{1.44} & \textbf{156.00 (0.00\%)} & \textbf{2.41} \\
				\bottomrule
		\end{tabular}}%
		\label{tbl:complexity}
	\end{center}
\end{table}

\section{Conclusion}

In this paper, we propose a module called FTNs for effectively and smoothly solving continuous-level learning problems in image processing. FTNs show very smooth results in practical applications because of their large filter similarity, producing reasonable performance on continuous levels. FTNs have fewer undesirable artifacts, more interpretability, and are extremely lightweight compared to the existing \textit{network tuning and interpolation} frameworks. We hope that our analysis of continuous-level learning and experiments in various scenarios can help in the development of real-world imaging applications.

{
	\bibliographystyle{aaai}
	\bibliography{egbib}
}

\end{document}